\documentclass[11pt]{article}

\usepackage[preprint]{acl}

\usepackage{times}
\usepackage{latexsym}

\usepackage[T1]{fontenc}

\usepackage[utf8]{inputenc}

\usepackage{microtype}

\usepackage{inconsolata}

\usepackage{graphicx}

\usepackage{amsmath}
\usepackage{amssymb}
\usepackage{mathtools}
\usepackage{amsthm}

\usepackage{hyperref}       
\usepackage{url}            
\usepackage{booktabs}       
\usepackage{amsfonts}       
\usepackage{amsmath}
\usepackage{nicefrac}       
\usepackage{microtype}      
\usepackage{xcolor}         
\usepackage{minted}
\usepackage{enumerate}
\usepackage{enumitem}
\usepackage{multirow}
\usepackage{algorithm}
\usepackage{algorithmic}

\usepackage[capitalize,noabbrev]{cleveref}

\title{From Atomic Actions to Standard Operating Procedures:\\ Iterative Tool Optimization for Self-Evolving LLM Agents}

\author{
 \textbf{Haipeng Ding\textsuperscript{1,2}},
 \textbf{Yuexiang Xie\textsuperscript{2}},
 \textbf{Zhewei Wei\textsuperscript{1,}\thanks{Correspondence to: Zhewei Wei $\langle$\href{mailto:zhewei@ruc.edu.cn}{zhewei@ruc.edu.cn}$\rangle$ and Yaliang Li$\langle$\href{mailto:yaliang.li@alibaba-inc.com}{yaliang.li@alibaba-inc.com}$\rangle$}},
 \textbf{Yaliang Li\textsuperscript{2,$\ast$}},
 \textbf{Bolin Ding\textsuperscript{2}}
\\
 \textsuperscript{1}Renmin University of China,
 \textsuperscript{2}Alibaba Group
\\
}

\begin{document}
\maketitle

\begin{abstract}
Tool utilization enables Large Language Model (LLM) agents to interact with the real world and resolve complex tasks.
However, existing agent frameworks predominantly rely on static toolsets composed of granular atomic actions (e.g., basic file I/O or single-turn search), which forces agents to reinvent low-level logic for every recurring workflow, leading to increased reasoning overhead and failure rates.
In this study, we propose that agents can achieve self-evolution by synthesizing these atomic actions into reusable Standard Operating Procedures (SOPs), which function as callable higher-order tools that encapsulate multi-step logic.
We further introduce \textsc{EvoSOP}, a framework that empowers agents to extract SOPs from execution trajectories and iteratively optimize the toolset through a systematic lifecycle of construction, merging, evaluation, and pruning. 
Extensive experiments demonstrate that \textsc{EvoSOP} significantly boosts task success rates while substantially reducing the number of interaction rounds compared to baselines. 
Our analysis also reveals that iterative tool optimization fosters reliable and efficient tool-use patterns, providing a scalable pathway for the development of self-evolving agents.
\end{abstract}

\section{Introduction}
\label{sec:introduction}
Large Language Models (LLMs)~\cite{DBLP:t5, DBLP:llama, DBLP:gpt3} have fundamentally advanced the field of artificial intelligence, demonstrating remarkable capabilities in logical reasoning and general-purpose problem-solving~\cite{DBLP:humaneval}. By leveraging external tools, LLM-based agents~\cite{DBLP:react, DBLP:dfsdt_toolllm, DBLP:codeact} extend these strengths beyond text generation to interact with the real world and solve complex and practical tasks~\cite{DBLP:gaiabench}.

While effective tool utilization is critical to the performance of agent systems, existing frameworks predominantly rely on static toolsets of atomic actions, such as basic file I/O, single-turn search, etc. 
This design forces agents to orchestrate every task through fine-grained sequences of low-level logic, diverging from the hierarchical efficiency seen in human problem-solving~\cite{DBLP:fastslowthink1}.
In practice, humans often bypass exhaustive deliberation by employing Standard Operating Procedures (SOPs) that encapsulate multi-step logic into cohesive and high-level routines. 
Without these abstractions, agents face significantly increased reasoning overhead and a higher risk of cascading errors, particularly in long-term tasks~\cite{DBLP:longhorizon1}.

Recent research on tool-augmented agents followed mainly two paths, refining a model's ability to use specific tools~\cite{DBLP:dfsdt_toolllm} and expanding the overall breadth of the available toolset~\cite{DBLP:gitagent}. 
However, these efforts do not address the fundamental inefficiency of reasoning in long and fragmented sequences of atomic actions. 
While recent studies~\cite{DBLP:craft, dblp:toolscope} enable agents to create new tools dynamically, they typically treat tool addition as a one-time event and lack a mechanism for long-term management. 
This leads to a bloated toolset where redundant or suboptimal tools accumulate, creating noise and complicating the agent's decision-making. 
This implies that a truly self-evolving agent requires more than just the ability to generate tools. 
It needs a systematic and iterative process to optimize its toolset by pruning ineffective tools, ensuring that evolved SOPs remain efficient and reliable.

To address the aforementioned challenges, we introduce \textsc{EvoSOP}, a framework that empowers agents to self-evolve by synthesizing atomic actions into high-level and reusable SOPs. 
Distinct from existing methods that treat tool creation as an isolated event, \textsc{EvoSOP} establishes a continuous optimization loop that progressively refines the agent's toolset. 
Specifically, the framework iteratively identifies recurring and useful patterns of atomic actions within execution trajectories, and compresses these long-horizon reasoning chains into callable higher-order tools. 
Within this same iterative cycle, \textsc{EvoSOP} governs the composition of the toolset by merging redundant routines and pruning low-utility actions based on their historical performance. This systematic approach ensures that agent capabilities remain both lean and powerful, effectively preventing performance degradation often caused by bloated toolsets. 
Note that \textsc{EvoSOP} serves as a model-agnostic framework that requires no parametric updates to the underlying LLMs, ensuring seamless integration with existing agent systems and enhancing the performance of black-box models.

Our main contributions are summarized as:
\begin{itemize}[leftmargin=0.2in, topsep=0in, itemsep=-0.048in]
\item We define a novel paradigm for agent self-evolving that formalizes how agent systems improve through iterative tool optimization. This paradigm establishes a conceptual parallel to the standard machine learning pipeline by mirroring essential stages such as data acquisition, forward execution, backward propagation, etc.
\item We propose \textsc{EvoSOP}, a framework that empowers agents to extract SOPs from their execution trajectories. The system iteratively optimizes the toolset through a systematic lifecycle that encompasses construction, merging, evaluation, and pruning.
\item We provide extensive experiments on various benchmarks to demonstrate the effectiveness of the proposed framework. The experimental results show that \textsc{EvoSOP} significantly increases task success rates while reducing the number of interaction rounds compared to baselines. 
\end{itemize}

\section{Preliminaries}
\label{sec:preliminaries}

We first introduce the definitions and notations of LLM-based agents. We focus on a practical agentic setting~\cite{DBLP:selfevolvesurvey, DBLP:agentreviewsurvey} where the agent interacts with an environment through a toolset to accomplish complex tasks. Due to the space limitation, we place another agent system named DFSDT~\cite{DBLP:dfsdt_toolllm} in Appendix~\ref{app:configs}.

\vspace{-0.05in}
\paragraph{Environment and Interaction.}
We consider an agent interacting with an environment $\mathcal{E}$. At each time step $t$, the agent receives an observation $o_t \in \mathcal{O}$ from the environment and takes an action $a_t \in \mathcal{A}$ based on its internal policy. The policy $\pi$ is instantiated by an LLM, expressed as:
\begin{equation*}
    a_t \sim \pi_{\text{LLM}}(\cdot \mid C_t),
\end{equation*}
where $C_t$ represents the \textit{context} available at time $t$. Following the paradigm of ReAct~\cite{DBLP:react}, the action $a_t$ may consist of a reasoning trace $l_t$ and a specific tool call $f_t$. 
For simplification, we denote the reasoning of the LLM given prompt $P$ as $a = \pi_{\text{LLM}}(P)$.
The environment then returns a response $r_t$ (e.g., execution results or error messages), which leads to the next state and observation $o_{t+1}$.

\vspace{-0.06in}
\paragraph{Context.}
The context $C_t$ is a structured representation of the agent's current state, designed to provide the LLM with all necessary information for decision-making. We formalize $C_t$ as a tuple:
\vspace{-0.06in}
\begin{equation*}
    C_t = \langle \mathcal{P}, \mathcal{G}, \mathcal{H}_t, \mathcal{M}_t, \mathcal{F} \rangle,
\end{equation*}
where $\mathcal{P}$ is the persona or system prompt defining the agent's role and behavioral constraints, $\mathcal{G}$ is the task goal and its associated success criteria, $\mathcal{H}_t = \{(a_1, r_1, o_1), \dots, (a_{t-1}, r_{t-1}, o_{t-1})\}$ is the interaction history, $\mathcal{M}_t$ represents the external memory or long-term knowledge retrieved by the agent, and $\mathcal{F}$ is the toolset containing the functional interfaces available to the agent.

\vspace{-0.06in}
\paragraph{Atomic Actions and Tool Schema.}
The toolset $\mathcal{F} = \{f_1, f_2, \dots, f_n\}$ initially consists of \textit{atomic actions}, i.e., granular and single-purpose functions such as file operations or API calls. Each tool $f_i \in \mathcal{F}$ is defined by its $\textit{schema} = \langle \text{name}, \text{desc}, \text{params}, \text{returns} \rangle$, where $\text{name}$ and $\text{desc}$ provide the semantic information required by the LLM to understand the tool's utility.

\vspace{-0.06in}
\paragraph{Workflow.}
The operational cycle of the agent follows a reasoning-acting loop. The agent is initialized with state $\mathcal{H}_0 = \{\}$. At each time step $t$, the agent performs the following operations:
\vspace{-0.06in}
\begin{align*}
    \text{Context Construction}&: P_t = \text{Format}(C_t),\\
    \text{Reasoning}&: \hat{a}_t = \pi_{\text{LLM}}(P_t), \\
    \text{Acting}&: r_t = \text{Exec}(\hat{a}_t, \mathcal{E}), \\
    \text{State Update}&: \mathcal{H}_{t+1} \leftarrow \mathcal{H}_t \cup \{(\hat{a}_t, r_t, o_t)\}.
\end{align*}
The process ends when the agent reaches a terminal state or exceeds the maximum iterations. 
We define an \textit{execution trajectory} $\xi_q=((\hat{a}_1, r_1, o_1), (\hat{a}_2, r_2, o_2), \cdots, (\hat{a}_k, r_k, o_k))$ of task $q$ when the agent is terminated after $k$ iterations.

\section{Methodology}

\begin{figure*}[t]
  \centering
  \includegraphics[width=0.96\textwidth]{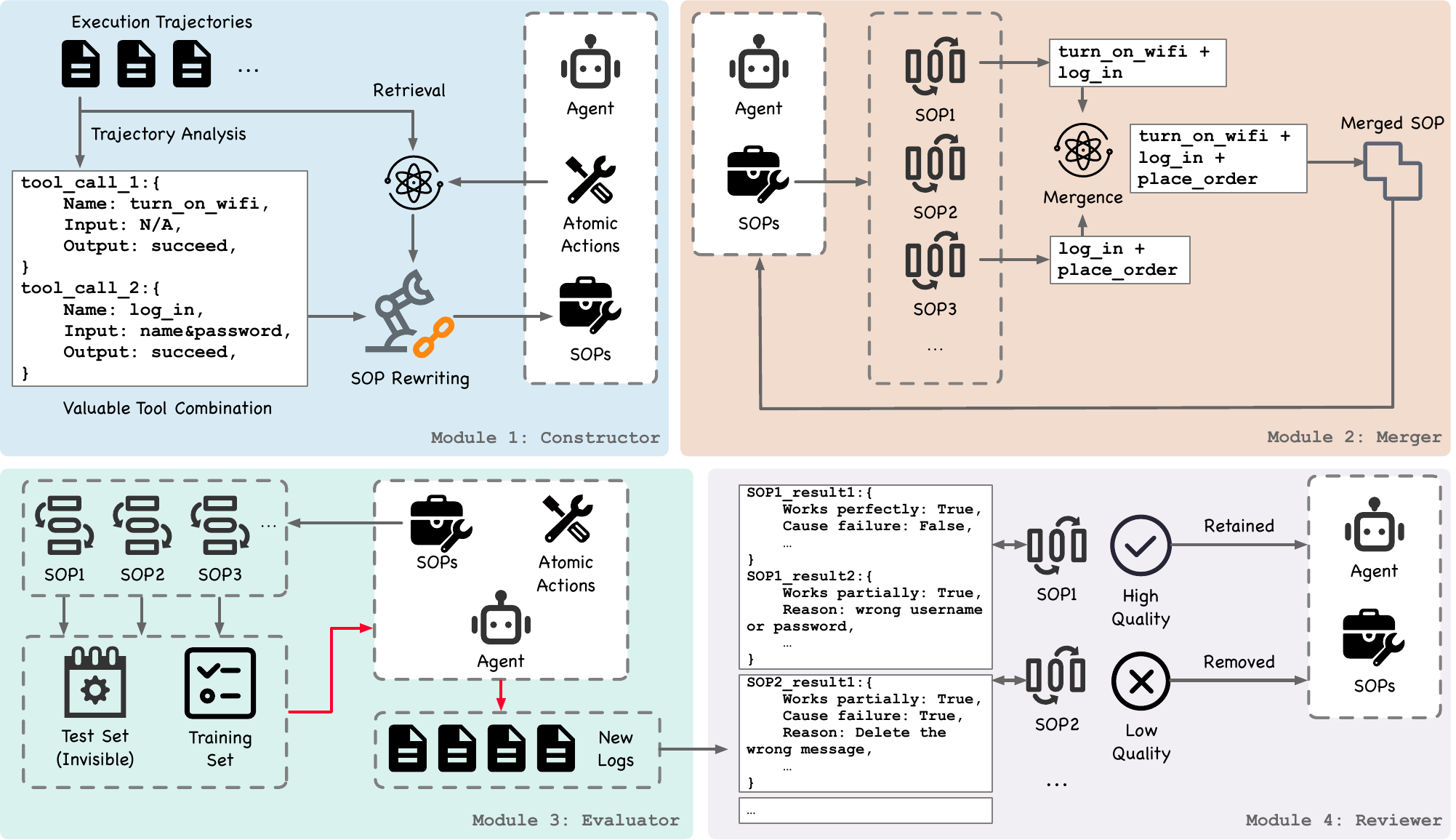}
  \vspace{-0.08in}
  \caption{The overall architecture of \textsc{EvoSOP}, illustrating the iterative tool optimization lifecycle. The framework employs four collaborative modules, including \textsc{Constructor}, \textsc{Merger}, \textsc{Evaluator}, and \textsc{Reviewer}.}
  \label{fig:init_overview}
  \vspace{-1.5em}
\end{figure*}


\subsection{Design Motivation and Principle}
\label{subsec:motivation}

As mentioned in Section~\ref{sec:introduction}, most LLM-based agents rely on a static toolset $\mathcal{F}$ composed of granular atomic actions. 
While recent studies have explored expanding the toolset through dynamic code generation~\cite{DBLP:asi, DBLP:craft}, they typically treat tool addition as a one-time event and lack a long-term management mechanism, leading to several critical limitations:

(i) \textit{Unreliable Tool Executability}: Although LLMs can generate syntactically correct code, synthesized tools often fail in complex environments due to various engineering reasons. Without a systematic evaluation and refinement, tools remain fragile and cannot be trusted for long-term tasks.

(ii) \textit{Limited Generalization and Reusability}: Tools derived from narrow scenarios often encode context-dependent logic that lacks broader applicability. Without a mechanism to iteratively abstract actions into reusable SOPs, the agent reinvents low-level logic, failing to achieve the hierarchical reasoning efficiency seen in human problem-solving.

(iii) \textit{Redundancy}: In the absence of a lifecycle management (e.g., merging and pruning), continuous tool creation leads to a bloated toolset and brings noise to the agent's context. Increasing reasoning overhead complicates the LLM's decision-making, ultimately degrading the success rate.

To address these challenges, \textsc{EvoSOP} moves beyond simple tool creation by establishing a continuous and iterative tool optimization process.

\subsection{Overview}
\label{subsec:framework_overview}

As illustrated in Figure~\ref{fig:init_overview}, \textsc{EvoSOP} consists of four collaborative modules that govern the optimization of the toolset, including:
(i) \textsc{Constructor}: This module identifies recurring action sequences within execution trajectories $\xi$, which extracts logical segments of atomic actions that frequently co-occur and synthesizes them into callable SOPs, complete with executable code and semantic schemas;
(ii) \textsc{Merger}: To maintain a lean toolset, this module inspects newly generated SOPs for functional redundancy, merging overlapping routines into generalized and higher-order tools to prevent the bloating of the context $C$;
(iii) \textsc{Evaluator}: The agent is equipped with the updated toolset $\mathcal{F}'$ and re-executes tasks. This module produces new trajectories that reflect the actual utility and reliability of the evolved SOPs in environments;
(iv) \textsc{Reviewer}: This module acts as a critic by analyzing the performance metrics of synthesized tools, which prunes SOPs that exhibit high error rates or significant redundancy compared to existing tools.


\vspace{-0.07in}
\paragraph{Agent Self-Evolution by Non-Parametric Learning.}
Although \textsc{EvoSOP} does not modify the learnable parameters of the underlying LLM, its iterative structure constitutes a form of non-parametric structural optimization. We formalize this by drawing a conceptual parallel between \textsc{EvoSOP} and the machine learning pipeline.

First, the agent's interaction with training tasks represents the \textit{data acquisition} stage, while the execution of tasks using the current toolset $\mathcal{F}$ corresponds to \textit{forward propagation}. The resulting execution trajectories $\xi$ serve as observable behavioral outputs, capturing the agent's reasoning process and tool-use patterns. Improvement is then achieved through a \textit{backward propagation}, where \textsc{EvoSOP} analyzes reasoning inefficiencies or execution failures within the trajectories. Instead of updating model weights via gradients, the framework performs ``symbolic'' backpropagation by extracting and synthesizing reusable SOPs that encapsulate multi-step logic into cohesive routines. 

Besides, to ensure the SOPs remain lean and to avoid the performance degradation typically caused by bloated toolsets, the merging and pruning processes function as critical \textit{regularization} mechanisms. These mechanisms mitigate the risk of ``overfitting'' to narrow and task-specific contexts by eliminating redundant or low-utility actions, thereby fostering generalized and reliable tool-use patterns. As the iteration proceeds, the stabilization of the toolset mirrors the behavior of a \textit{decaying learning rate}, where the agent eventually converges toward an optimized hierarchy of toolset. 

\subsection{Tool Optimization Lifecycle in \textsc{EvoSOP}}
\label{sec:mainidea}

In this subsection, we detail the tool optimization lifecycle in \textsc{EvoSOP}, which is designed to systematically transform raw interaction experiences into a solidified toolset $\mathcal{F}$ of high-level SOPs.
Please refer to Appendix~\ref{app:prompts} for the selected prompts adopted in \textsc{EvoSOP} indicating our design conception.




\vspace{-0.05in}
\paragraph{\textsc{Constructor}: From Trajectories to Functional Abstractions.} 
The optimization lifecycle begins with the \textsc{Constructor}, an LLM-based module designed to distill reusable functional abstractions from historical execution trajectories $\xi$. As established in Section~\ref{sec:preliminaries}, raw trajectories record sequences of atomic actions and environment responses in strict temporal order. In practice, many of these consecutive tool calls are not merely coincidental but reflect deep-seated logical or causal dependencies required to resolve recurring sub-problems. The \textsc{Constructor} identifies these patterns to transform fragmented action sequences into cohesive and higher-order routines.

The primary objective of this stage is to mitigate the excessive reasoning overhead inherent in static and granular toolsets. \textsc{Constructor} extracts empirically valuable segments of tool-use patterns and functionalizes them into callable code. These synthesized SOPs are more than simple linear macros; they are interleaved with lightweight processing logic (e.g., conditional checks or error handling) to ensure they remain applicable across varying environmental states (an implementation example is provided in Appendix~\ref{code:sop_example}).

We formalize the SOP construction process as a two-stage transformation:
\begin{equation*}
    \mathcal{\xi}'_i = f_{\text{extract}}(\mathcal{\xi}_i), \quad
    \mathcal{S}_i = f_{\text{rewrite}}(\mathcal{\xi}'_i, \mathcal{F}_{\text{atomic}}),
\end{equation*}
where $\mathcal{\xi}_i$ represents the raw execution trajectory for task $i$, and $\mathcal{\xi}'_i$ denotes the set of action segments identified as logically coupled. The transformation function $f_{\text{rewrite}}$ then maps these segments into the resulting SOP source code $\mathcal{S}_i$, drawing upon the functional interfaces of the initial atomic toolset.

\vspace{-0.05in}
\paragraph{\textsc{Merger}: Structural Optimization.}
As the agent accumulates experience across task domains, functional overlap and redundancy among synthesized SOPs become inevitable, often manifesting as noise within the context $C_t$. Such noise complicates the LLM's decision-making and may lead to the selection of suboptimal or conflicting tools. To address this, we design the \textsc{Merger}, a module dedicated to simplifying the toolset and enhancing its overall quality through structural optimization.

The \textsc{Merger} analyzes the candidate SOP set within a training batch to identify functional overlaps based on shared objectives or highly similar logic. When candidates with convergent functionalities are identified, the module integrates them into a single, more expressive SOP that preserves the constituent capabilities of its predecessors while maintaining a high degree of generality. We formalize this batch-based merging process as follows:
\begin{equation*}
    \mathcal{S}_b = \bigcup_{i\in b}\mathcal{S}_i, \quad
    \mathcal{S}' = f_{\text{merge}}(\mathcal{S}_b),
\end{equation*}
where $b$ denotes the set of task indices, $\mathcal{S}_b$ is the union of newly constructed SOPs, and $\mathcal{S}'$ represents the consolidated set generated by the merger.

\textsc{EvoSOP} adopts a non-destructive consolidation strategy during this stage. While newly merged, composite SOPs are added to the toolset, the original constituent SOPs are not immediately removed. This design choice is motivated by two critical observations regarding agent self-evolution. Firstly, even when two SOPs possess nominally identical functionality, they may differ significantly in implementation quality, including robustness, coding style, and the clarity of their docstrings, all of which affect the LLM's ability to invoke them correctly. Secondly, a larger composite SOP is not inherently superior to its more granular components; it may introduce implicit technical defects or reduce execution reliability in specific contexts. By deferring the removal of tools until the verification phase, \textsc{EvoSOP} ensures that the final toolset is refined based on empirical performance rather than static heuristics, adhering to the principle of hierarchical efficiency without sacrificing reliability.



\vspace{-0.05in}
\paragraph{\textsc{Evaluator}: Execution-based Validation.}
As synthesized SOPs are inherently prone to fragility when deployed in complex and stateful environments, they cannot be immediately integrated into the agent's core capabilities without rigorous validation. \textsc{EvoSOP} subjects all candidate SOPs to an \textsc{Evaluator}, which serves as the experimental foundation for assessing their real-world utility.

Specifically, all constructed SOPs within the sets $\mathcal{S}_b$ and $\mathcal{S}'$ are loaded into the execution environment. Since these SOPs are instantiated as callable functions, the framework must first bridge the gap between executable code and the LLM's reasoning interface. We accomplish this by extracting the functional schema for each SOP (i.e., comprising its name, semantic description, and parameter specifications) and updating the toolset $\mathcal{F}$ as:
\begin{equation*}
    \mathcal{F}' = \mathcal{F_\text{atomic}} \cup \left\{ f_{\text{schema}}(s) \mid s \in \mathcal{S}_b \cup \mathcal{S}' \right\}.
\end{equation*}
This expanded toolset $\mathcal{F}'$ provides the agent with a hierarchical choice. It can either continue using granular atomic actions or invoke the newly synthesized SOPs to bypass multi-step reasoning.

Following the toolset update, \textsc{EvoSOP} initiates a full re-execution of the training tasks within a real-world setting. During this process, the framework meticulously monitors task outcomes and captures the resulting execution trajectories $\hat{\xi}$. These trajectories provide the empirical data necessary to diagnose implementation defects, assess functionality under diverse conditions, and provide a principled basis for the final quality control process.

\vspace{-0.05in}
\paragraph{\textsc{Reviewer}: Quality Control and Simplification.}
The final stage of the \textsc{EvoSOP} lifecycle is a rigorous quality control procedure designed to ensure the reliability and effectiveness of the toolset. As SOPs are synthesized independently across varied trajectories, several systematic issues typically arise: (i) functional redundancy among overlapping tools; (ii) semantic misalignment, where misleading tool names or docstrings trigger retrieval errors; (iii) limited utility of tools that are rarely invoked; and (iv) latent technical defects that manifest only in specific stateful contexts. Addressing these challenges is essential to prevent the accumulation of technical debt and the resulting performance degradation caused by bloated toolsets.

Within the reasoning-acting loop, an agent's understanding of an SOP is governed by its docstring. Since these SOPs are automatically generated from historical logs, any discrepancy between the tool's implementation and its semantic description can lead to inappropriate invocations or reasoning failures. To operationalize the assessment of such risks, we introduce \textsc{Reviewer}, an LLM-based module that functions as a critic. The \textsc{Reviewer} analyzes the verification trajectories $\hat{\xi}$ and categorizes each SOP invocation into the following states:
\begin{itemize}[leftmargin=0.2in, topsep=0in, itemsep=-0.048in]
    \item \textit{Optimal Execution}: The SOP successfully completes its intended functionality as defined in its docstring without any technical exceptions.
    \item \textit{Partial Utility}: The tool executes without technical error but achieves its intended goals partially with complicated reasons.
    \item \textit{Neutrality}: The SOP returns successfully but produces no significant change in the environmental state, indicating low practical relevance.
    \item \textit{Negative Interference}: The tool negatively impacts task progress or compromises the environmental state, leading to unacceptable outcomes.
    \item \textit{Implementation Defect}: Execution triggers a technical exception and produces a traceback, indicating internal implementation errors.
\end{itemize}

These statuses are not mutually exclusive, reflecting the nuanced failure modes of complex agentic systems. For each judgment, the \textsc{Reviewer} provides a brief natural-language justification, ensuring that the pruning process is both interpretable and traceable. Following the aggregation of these performance statistics across the training batch, \textsc{EvoSOP} performs a global filtering operation to refine the toolset:
\begin{align*}
    \mathcal{R} &= \bigoplus_{i \in I_{\text{train}}} f_{\text{review}}(\hat{\xi}_i), \nonumber\\
    \mathcal{\mathcal{S}}_{\text{solid}} &= \{ s \mid f_{\text{check}}(s, \mathcal{R}_s) \neq \text{remove}, s \in \mathcal{S}' \cup \mathcal{S}_b \},
\end{align*}
where $\oplus$ is the review aggregation operator. 


\subsection{Training Workflow}
As formally specified in Section~\ref{sec:mainidea}, each component of \textsc{EvoSOP} is designed to support a parameter-free training paradigm. Unlike prior studies that predominantly rely on parametric fine-tuning, \textsc{EvoSOP} realizes the ``training'' process through iterative optimization of the toolset. The complete training workflow is presented in Algorithm~\ref{alg:EvoSOP} in Appendix~\ref{app:pseudo-code}.




\vspace{-0.05in}
\paragraph{Mini-batching.}
At the beginning of the training process, we partition the initial execution trajectories $\mathcal{\xi}$ into discrete mini-batches. In those batches, the \textsc{Constructor} identifies tool-use patterns and proposes new functional abstractions. Notably, this is the exclusive stage for introducing new SOPs, ensuring the growth of toolset is strictly grounded in observed execution experience.

In every iteration, the \textsc{Merger} first consolidates overlapping functionalities into generalized procedures to maintain a compact toolset. Subsequently, the system enters an evaluation phase where the agent re-executes all training tasks with the updated SOPs. Using the resulting rollouts, the \textsc{Reviewer} assesses each tool's empirical contribution and prunes those that exhibit implementation defects or redundant reasoning. 

\vspace{-0.05in}
\paragraph{Checkpointing.}
A critical challenge in LLM-based optimization is the potential stochasticity of model behavior, where aggressive pruning might inadvertently remove genuinely beneficial SOPs. To mitigate this risk, \textsc{EvoSOP} avoids a strictly linear update; instead, it adopts a checkpointing mechanism. At the end of each iteration, the full evaluation logs and the current SOP toolset are archived. Since task feedback is obtained directly from the environment in a self-supervised manner, we do not require a separate validation set. We select the iteration that yields the highest training success rate as the final epoch, ensuring that the output toolset $\mathcal{S}^*$ represents the optimal balance between expressivity and reliability. 

\section{Experiments}

\subsection{Setup}
Our evaluation compares several distinct tool configurations: (i) {\bf ReAct} and {\bf DFSDT}, which use the original atomic actions; (ii) {\bf ASI}~\cite{DBLP:asi}, whose toolset is augmented with SOPs, referred to as {\it skills} in the original paper, induced through a one-time process; (iii) {\bf DRAFT}~\cite{DBLP:draft}, which equips tools with rewritten descriptions and usage guidance; and (iv) {\bf EvoSOP}, our proposed method that performs iterative SOP optimization. The evaluation is conducted on ACEBench~\cite{DBLP:acebench} and Tau2Bench~\cite{DBLP:tau2bench}. For ACEBench, we use the agent subset, while for Tau2Bench, we use the solo mode of the Telecom subset.

Throughout the experiments, \texttt{GPT-4o} serves as the backbone LLM for all evaluation agents. During the toolset construction phase, we adopt \texttt{GPT-4o}, \texttt{Gemini-3-Flash-Preview}, and \texttt{Qwen-Max} to assess cross-model robustness. Detailed configurations are provided in Appendix~\ref{app:configs}.


\subsection{Performance Comparisons}
The main experimental results are summarized in Table~\ref{tbl:res-main}. 
In general, \textsc{EvoSOP} outperforms both the base agent and other selected tool-related baselines. On both subset of ACEBench, \textsc{EvoSOP} achieves a significant improvement over the base method, yielding gains ranging from 2.5\% to 13.4\% depending on the backbone model. This performance surge indicates that the evolved SOPs effectively address recurring reasoning bottlenecks by providing the agent with robust and reliable logic blocks. 

\begin{table}[!t]
\caption{Averaged successful rate $\pm$ standard error (\%) of the baselines and \textsc{EvoSOP} on benchmark ACEBench and Tau2Bench. Note that all the base agentic methods relied by tool-related methods are run on GPT-4o.}
\vspace{-0.08in}
\label{tbl:res-main}
\centering
\resizebox{0.9\linewidth}{!}{
\begin{tabular}{lccc}
\toprule
\multicolumn{4}{c}{ACEBench Multi-Step} \\
Backbone & Gemini-3-FP & GPT-4o & Qwen-Max \\
\midrule
ReAct & / & $80.8\pm3.4$ & / \\
\ + ASI & $45.8\pm14.8$ &$60.0\pm12.9$ & $53.3\pm9.4$ \\
\ + EvoSOP & $\mathbf{83.3\pm2.4}$ & $\mathbf{84.2\pm1.9}$ & $\mathbf{84.2\pm1.9}$\\
DFSDT & / & $78.3\pm6.8$ & / \\
\ + DRAFT & $\mathbf{84.2\pm1.9}$ & $75.0\pm5.8$ & $74.2\pm5.4$\\
\ + EvoSOP & $82.5\pm2.6$ &$\mathbf{85.0\pm0.0}$ & $\mathbf{85.8\pm1.9}$\\
\midrule
\multicolumn{4}{c}{ACEBench Multi-Turn} \\
ReAct & / & $72.2\pm3.7$ & / \\
\ + ASI & $46.7\pm11.1$ & $57.2\pm8.0$ & $65.6\pm8.8$\\
\ + EvoSOP & $\mathbf{79.4\pm4.5}$ & $\mathbf{79.4\pm3.0}$ & $\mathbf{84.4\pm3.1}$\\
DFSDT & / & $69.4\pm4.5$ & / \\
\ + DRAFT & $71.1\pm6.0$& $66.1\pm3.0$&$\mathbf{78.3\pm2.6}$ \\
\ + EvoSOP & $\mathbf{83.3\pm3.3}$ &$\mathbf{74.4\pm3.1}$ & $77.9\pm4.1$\\
\midrule
\multicolumn{4}{c}{Tau2-Bench Telecom Solo}\\
ReAct & / & $37.1\pm2.8$ & / \\
\ + ASI & $35.4\pm1.1$ & $37.7\pm1.4$ & $39.5\pm4.0$\\
\ + EvoSOP & $\mathbf{43.3\pm0.8}$ & $\mathbf{40.9\pm0.4}$ & $\mathbf{40.2\pm2.2}$\\
\bottomrule

\end{tabular}
}
\vspace{-1em}
\end{table}

In the challenging Tau2Bench-Telecom, which is characterized by a larger tool space and intricate stateful transitions, \textsc{EvoSOP} maintains a steady performance lead. Although the margin of improvement is more concentrated compared to ACEBench due to the inherent complexity of the domain, the results show that \textsc{EvoSOP} can navigate and optimize tools even in high-entropy scenarios where static or one-shot toolsets typically struggle.

From the experiments, we also observe that \textsc{EvoSOP} produces reliable SOPs. Our analysis of the trajectories reveals that GPT-4o often encounters reasoning failures as the context window grows or when tasks require precise temporal tracking. For example, in text message management scenario (e.g., deleting the earliest message), agents using atomic actions often become indecisive after multiple turns, frequently returning control to the user for clarification, or deleting the wrong target. 

While ASI can induce SOPs for such workflows, its one-shot nature often leads to brittle logic that may delete the latest message instead of the earliest, causing cascading errors in similar but slightly different contexts. In contrast, the behavior of DRAFT becomes more stable as it does not make substantial changes to the tools. \textsc{EvoSOP} employs its \textsc{Reviewer} and \textsc{Merger} to evaluate execution feedback. It identifies these subtle logic flaws, merges redundant tools, and prunes SOPs that exhibit high error rates. This iterative self-correction and complete lifecycle management ensure that the resulting toolset is not just larger, but fundamentally more reliable and useful.

\begin{figure}[t]
  \centering
  \includegraphics[width=0.92\linewidth]{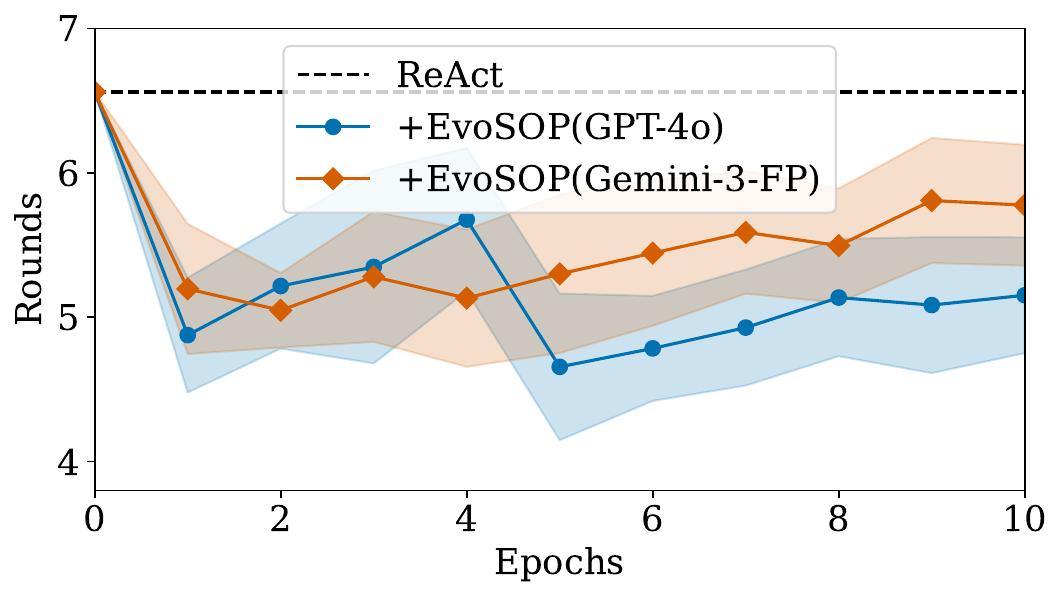}
  \vspace{-0.08in}
  \caption{The averaged reasoning rounds across epochs in dataset ACEBench .}
  \vspace{-0.2in}
  \label{fig:rounds_std}
\end{figure}

\vspace{-0.06in}
\paragraph{Reasoning Compression.}
A core hypothesis in this study is that SOPs should reduce the cognitive load on the agent by encapsulating multi-step workflows. In Figure~\ref{fig:rounds_std}, we illustrate the average number of reasoning rounds per task across the epochs. 

The experimental results confirm that the transition from atomic actions to SOPs substantially streamlines the interaction process. In the initial phases, the toolset may contain experimental or unstable SOPs, occasionally leading to redundant trial-and-error reasoning. However, as the iterative cycle progresses, \textsc{EvoSOP} filters out low-utility tools and consolidates overlapping routines. By the final several epochs, the number of interaction rounds stabilizes at a significantly lower level than that of baselines. Such reasoning compression not only reduces API latency and cost but also minimizes the risk of the agent losing focus within long-horizon trajectories, which is a primary driver of the observed increase in success rates.

\subsection{Ablation Study}

\begin{figure}[t]
  \centering
  \includegraphics[width=0.94\linewidth]{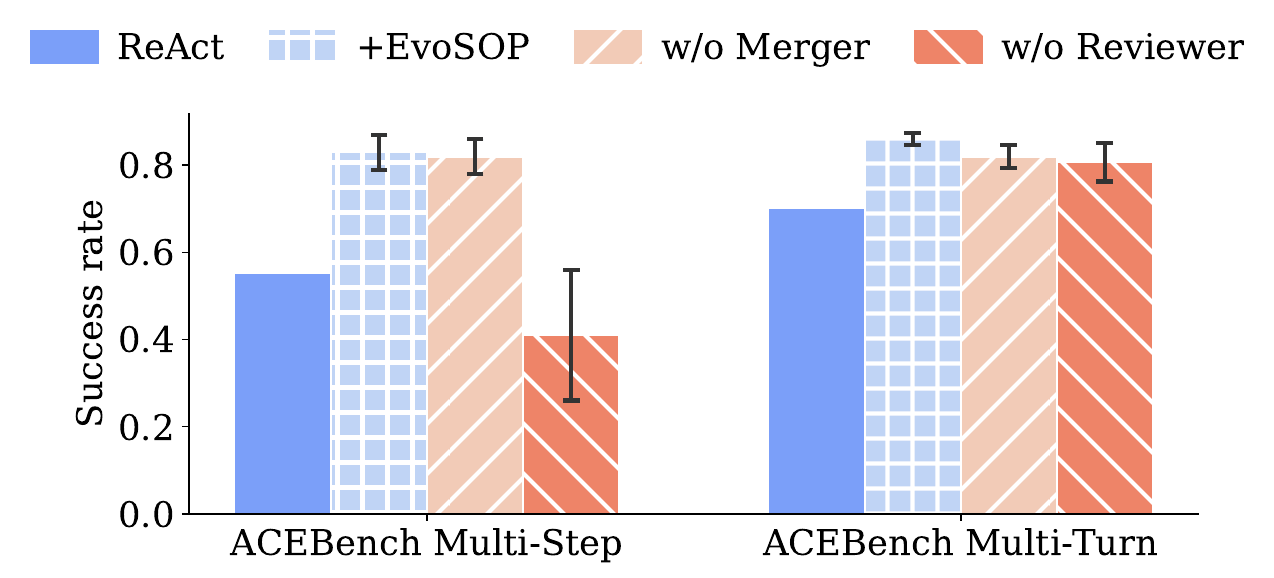}
  \vspace{-0.08in}
  \caption{Performance of \textsc{EvoSOP} and its ablated variants in dataset ACEBench.}
  \vspace{-0.2in}
  \label{fig:abres}
\end{figure}

To quantify the individual contribution of each component within the proposed \textsc{EvoSOP}, we conduct an ablation study by systematically removing the \textsc{Reviewer} and \textsc{Merger} modules. Figure~\ref{fig:abres} illustrates the average performance of the complete \textsc{EvoSOP} and its variants. 
Overall, the ablation study demonstrates that the superior performance of \textsc{EvoSOP} is not derived from any single module but from the synergy of its various components, which together provide comprehensive tool lifecycle management.

Specifically, the experimental results reveal that the iterative reviewing and pruning procedure, driven by the \textsc{Reviewer}, has the most profound impact on final performance. When the pruning mechanism is disabled, the success rate of the agent significantly declines on both datasets. These results confirm that, while the \textsc{Constructor} introduces beneficial SOPs, it also inevitably generates low-quality or context-sensitive tools that can introduce noise. 
Without the ``backward propagation'' of performance feedback provided by the \textsc{Reviewer}, these suboptimal tools accumulate, leading to a bloated toolset that confuses the agent’s decision-making. These findings explain the performance gap between \textsc{EvoSOP} and the one-shot ASI baseline (as shown in Table~\ref{tbl:res-main}), reinforcing that active tool management is just as critical as tool creation for self-evolving agents.

Besides, we evaluate the necessity of the \textsc{Merger} module. As shown in Figure~\ref{fig:abres}, the variant without the merging process also exhibits performance degradation. 
The \textsc{Merger} performs semantic and functional consolidation, transforming narrow and task-specific action sequences into generalized SOPs. By merging overlapping routines, the framework maintains a compact toolset where each SOP possesses a broader scope of applicability and higher reliability. These results indicate that these consolidated tools are invoked more frequently across diverse scenarios, which in turn reduces their risk of being erroneously evicted during the pruning phase. Consequently, the \textsc{Merger} acts as a ``regularizer'' that fosters the emergence of robust, high-order tool-use patterns, ensuring the long-term stability of the agent system.

\subsection{Case Study}

\begin{figure}[t]
  \centering
  \includegraphics[width=0.96\linewidth]{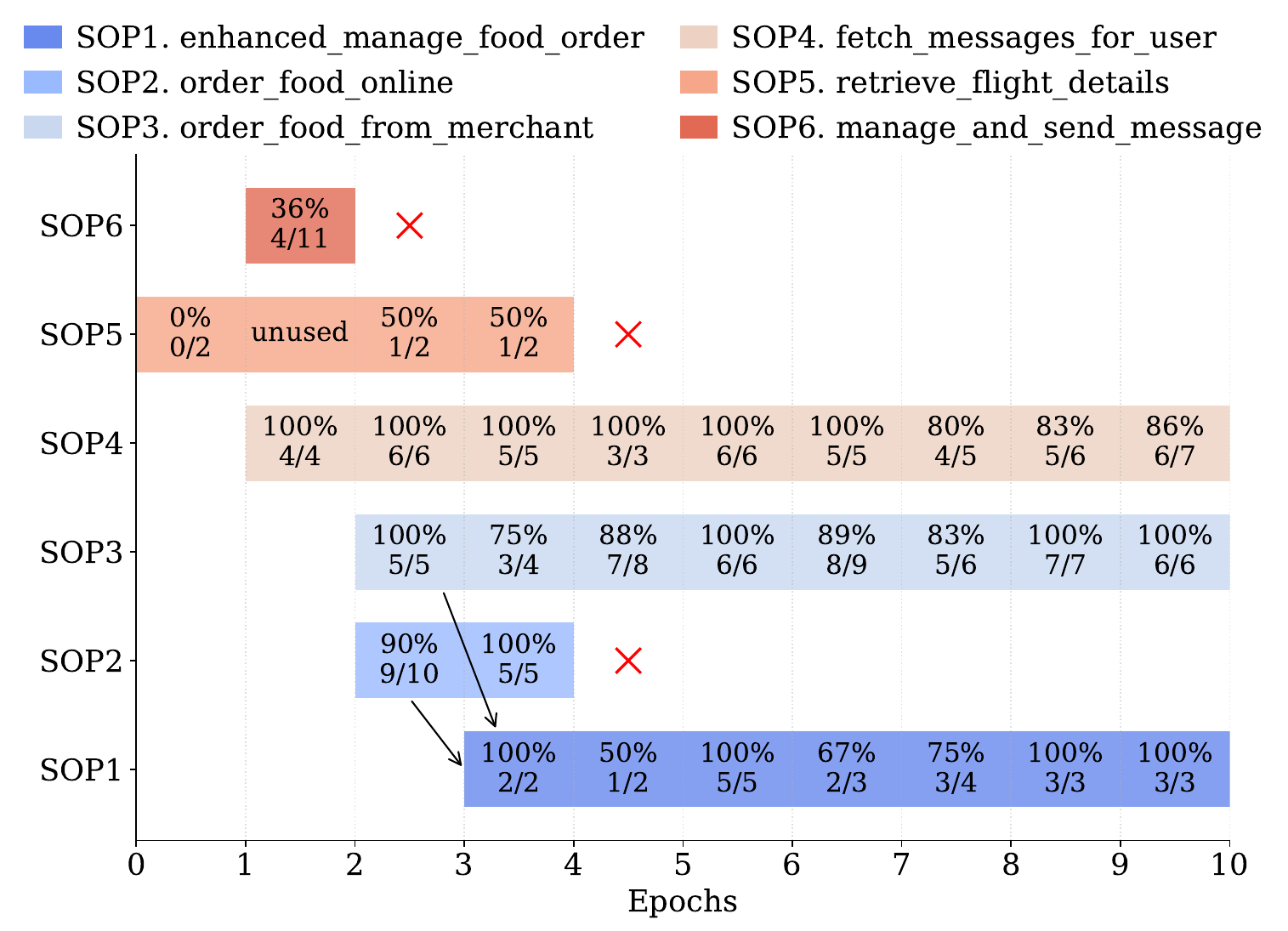}
  \vspace{-0.08in}
  \caption{The lifetime of synthesized SOPs. The numbers inside each block represent the success rate (\#success/\#total) of the SOP. Lines ending with a cross indicate that the corresponding SOPs have been removed during the optimization cycle.}
  \label{fig:SOPlifetime}
  \vspace{-0.2in}
\end{figure}

To better understand the optimization process, we analyze a representative execution of \textsc{EvoSOP} on ACEBench. 
We track the utilization and performance metrics of selected SOPs to illustrate how \textsc{EvoSOP} distinguishes between high-utility, high-risk, and redundant tools. As shown in Figure~\ref{fig:SOPlifetime}, we observe some distinct lifetime patterns:

(i) \textit{Fundamental SOPs}: Tools like \texttt{fetch\_ message} exhibit high invocation frequency and near-zero error propensity. These represent stable logic that the framework identifies early and retains as core components of agents.

(ii) \textit{Transient SOPs}: \texttt{manage\_and\_send\_mess- age} is synthesized during Epoch 1 but immediately evicted by the \textsc{Reviewer} due to a critical failure rate. This demonstrates the framework's ability to prevent unreliable code from polluting the toolset. Similarly, \texttt{retrieve\_flight\_details} is pruned after several epochs because its marginal utility is low, it is rarely invoked, and introduces non-negligible reasoning overhead without a proportional increase in success rate.

(iii) \textit{Merged SOPs}: Initially, the \textsc{Constructor} generated two task-specific tools, i.e., \texttt{order\_food\_online} and \texttt{order\_food\_from\_ merchant}. Later the \textsc{Merger} identified their functional overlap and synthesized a generalized SOP, \texttt{enhanced\_manage\_food\_order}. Consequently, the original specific tools are pruned to maintain toolset leanness, while the new SOP continued to perform reliably across diverse contexts.

This case clearly demonstrates how the designed components cooperate to produce a compact SOP toolset with high-quality SOPs. In general, the size of the SOP set remains consistently low, even as new SOPs are continually generated. We summarize the macro-level evolution of the toolset composition in Appendix~\ref{app:macrotrend}, showing that \textsc{EvoSOP} maintains the toolset at a compact scale.

\section{Related Works}


Early research primarily focuses on enhancing tool mastery through parametric updates, employing supervised fine-tuning or reinforcement learning to improve tool retrieval and invocation accuracy~\cite{DBLP:kimik2, DBLP:dfsdt_toolllm, DBLP:deepagent}. In contrast, non-parametric methods optimize tool use without modifying the underlying model. For example, DRAFT~\cite{DBLP:draft} and EasyTool~\cite{DBLP:easytool} refine tool documentation and retrieval strategies through trial feedback. 
Moving toward self-evolution, several studies empower agents to synthesize new tools via code generation, such as Voyager~\cite{arxiv:voyager}, CRAFT~\cite{DBLP:craft}, and Alita~\cite{DBLP:alita}, while ASI~\cite{DBLP:asi} introduces a mechanism to induce high-level skills from successful action trajectories. However, these works typically treat tool creation as a one-off event and lack a long-term management mechanism, often leading to toolkit bloating and reasoning redundancy, which motivates the development of~\textsc{EvoSOP} in this study.

\section{Conclusions}

In this study, we propose \textsc{EvoSOP}, a novel framework designed to empower LLM-based agents to self-evolve through iterative tool optimization. By synthesizing granular atomic actions into high-level and reusable SOPs, \textsc{EvoSOP} boosts task success
rates and addresses the reasoning overhead inherent in static toolsets. Different from existing methods that treat tool creation as a one-time event, our framework establishes a continuous optimization lifecycle, effectively mitigating the risks of logic fragmentation and toolset bloating. Extensive experiments on ACEBench and Tau2Bench demonstrate that \textsc{EvoSOP} significantly enhances task success rates while reducing the number of reasoning rounds across diverse benchmarks. 

\clearpage
\bibliography{custom}
\appendix
\newpage




\section{Pseudo-code of Training Workflow}
\label{app:pseudo-code}
In this section, we present the pseudo-code to clearly illustrate the main training workflow of our method. While the descriptions in the main text are module-centric, the pseudo-code provides a global overview; consequently, some notational differences may exist between the two. The correspondences are as follows:

\begin{itemize}
\item The function $f_\text{merge}$ takes as input the union of the SOP set maintained from previous iterations and the SOPs newly generated in the current iteration.
\item $B_i$ denotes the $i$-th mini-batch in the index set $B$, corresponding to $b$.
\item $S$ represents the dynamically generated and pruned SOP set, which eventually becomes the solidified SOP set $S_\text{solid}$.
\end{itemize}

\begin{algorithm}[tb]
    \caption{EvoSOP training workflow}
    \label{alg:EvoSOP}
    \begin{algorithmic}
        \STATE {\bfseries Input:} execution log set $\mathcal{\xi}$, atomic tool set $\mathcal{F}$, training set index $I_{\mathrm{train}}$, batched index set $B$ 
        \STATE {\bfseries Built in:} max iterations $M$, LLM-based function $f_{\cdot}(\cdot)$, verification module $g_\mathrm{verify}(\cdot)$
        \STATE Initialize $\mathcal{S}=\phi,\mathcal{R}=\phi$
        \FOR{$i=0$ {\bfseries to} $M-1$}
            \IF{$i<\mathrm{len}(B)$}
                \STATE $\mathcal{S}_{B_i}=\phi$
                \FOR{$j$ {\bfseries in} $B_i$}
                    \STATE$\mathcal{S}_{B_i}=\mathcal{S}_{B_i}\cup f_{\mathrm{rewrite}}(f_{\mathrm{extract}}(\mathcal{\xi}_j),\mathcal{F})$
                \ENDFOR
            \ENDIF
            \STATE $\mathcal{S}=\mathcal{S}\cup \mathcal{S}_{B_i}\cup f_{\mathrm{merge}}(\mathcal{S}\cup\mathcal{S}_{B_i})$
            \STATE $\mathcal{F}'= \mathcal{F}\cup\left\{f_\mathrm{schema}(s)|s\in S\right\}$
            \STATE $\widehat{\mathcal{\xi}}=g_\mathrm{verify}(\mathcal{F}')$
            \FOR{$j$ {\bfseries in} $I_\mathrm{train}$}
                \STATE $\mathcal{R}=\mathcal{R}\oplus f_\mathrm{review}(\widehat{\mathcal{\xi}_i},\mathcal{S})$
            \ENDFOR
            \FOR{$s$ {\bfseries in} $\mathcal{S}$}
                \IF {$\mathcal{R}_s\ \mathrm{exists}\ \mathrm{and}\  f_\mathrm{check}(\mathcal{R}_s)$==``remove''}
                    \STATE $\mathcal{R}=\mathcal{R}\setminus\mathcal{R}_s$
                    \STATE $\mathcal{S}=\mathcal{S}\setminus s$
                \ENDIF
            \ENDFOR
            \STATE save $\mathcal{S},\mathcal{R}$
        \ENDFOR
    \end{algorithmic}
\end{algorithm}

\section{Detailed Experimental Settings and Configurations}
\label{app:configs}

In this section, we provide a comprehensive description of the experimental configurations, settings, and implementation details. This section offers a detailed overview of the parameters and environments used in our study, ensuring the transparency and reproducibility of the reported results.

\paragraph{Dataset information.} In our experiments, we evaluate our method and the baselines on ACEBench and Tau2Bench. ACEBench is a tool-use-oriented benchmark, in which the \texttt{Agent-Multi-Step} and \texttt{Agent-Multi-Turn} subsets focus on evaluating agents' task-completion ability when equipped with different tools. Tau2Bench is a benchmark for conversational AI agents that involves issue resolution and simulated human feedback. To align with the pure tool-use setting considered in our study, we use the solo mode of the Telecom subset, which focuses on resolving phone network connection issues by the agent itself.

\paragraph{Data preparation for EvoSOP and baselines.} For each dataset, we first run the base methods, namely ReAct and DFSDT, to collect an initial set of trajectories, regardless of whether the trajectories are successful. During the training process of EvoSOP, we feed 5 randomly sampled trajectories in each of the first 5 iterations, exposing our method to 25 trajectories and tasks in total. In the last 5 iterations, EvoSOP only performs merging, evaluation, and review, meaning that no new SOPs are generated from trajectories. 

To ensure a fair comparison, we expose the same number of trajectories to the ASI baseline for agentic skill induction. The reported results include performance on both exposed and unseen tasks.

\paragraph{Baselines.} In addition to the ReAct agentic system described earlier, we reproduce another well-known method, DFSDT~\cite{DBLP:dfsdt_toolllm}, which implements a DFS-based algorithm to explore potentially promising choices and backtrack from failed branches.

DFSDT allows the agent to give up and backtrack to a previous state when encountering obstacles during task solving. Typically, when \texttt{max\_beam\_size} is set to 1, DFSDT effectively degenerates into a ReAct agent with the additional option of giving up. DFSDT dumps and restores the current states of both the agent and the environment before each reasoning step. When DFSDT gives up on a search branch, it restores the working state from the previously saved backup and performs a re-reasoning step, providing information about the failed branches in the prompt so that the agent can avoid repeating the same mistakes. For both ReAct and DFSDT, we set \texttt{max\_steps} to 100, and we set \texttt{max\_beam\_size} of DFSDT to 3.

For the two tool-related methods, {\bf ASI} and {\bf DRAFT}, we select them as representative approaches for one-shot high-level tool optimization and tool-docstring optimization, respectively. We apply them to both ReAct and DFSDT for comparison. For all methods mentioned above, we run 6 complete workflows on ACEBench and 3 complete workflows on Tau2Bench. The results reported in Table~\ref{tbl:res-main} are the average success rates $\pm$ standard errors across runs.

All of the used artifacts including benchmarks and code bases are consistent with their intended use.

\paragraph{Backbone models.} To ensure a fair comparison within our experiments, we consistently use \texttt{GPT-4o} as the backbone model for executing vanilla ReAct and DFSDT, as well as for the evaluation and testing phases of ASI, DRAFT, and EvoSOP, since these phases depend on the base agent methods. This is why we do not report the results of ReAct and DFSDT with \texttt{Gemini-3-Flash-Preview} or \texttt{Qwen-Max} as the backbone model. For the core algorithms of ASI, DRAFT, and EvoSOP, we adopt \texttt{GPT-4o}, \texttt{Gemini-3-Flash-Preview}, and \texttt{Qwen-Max} to assess the cross-model robustness. 
Detailed documentation regarding the model's training data coverage, linguistic capabilities, and known limitations can be found in the official technical report.

Our proposed method is parameter-free. Therefore, we did not perform any local model training. The computational cost is limited to the inference time via API calls to the proposed models. All experiments were conducted on a standard CPU.

\section{Toolset Maintenance and Convergence.}
\label{app:macrotrend}
\begin{figure}[t]
  \centering
  \includegraphics[width=0.9\linewidth]{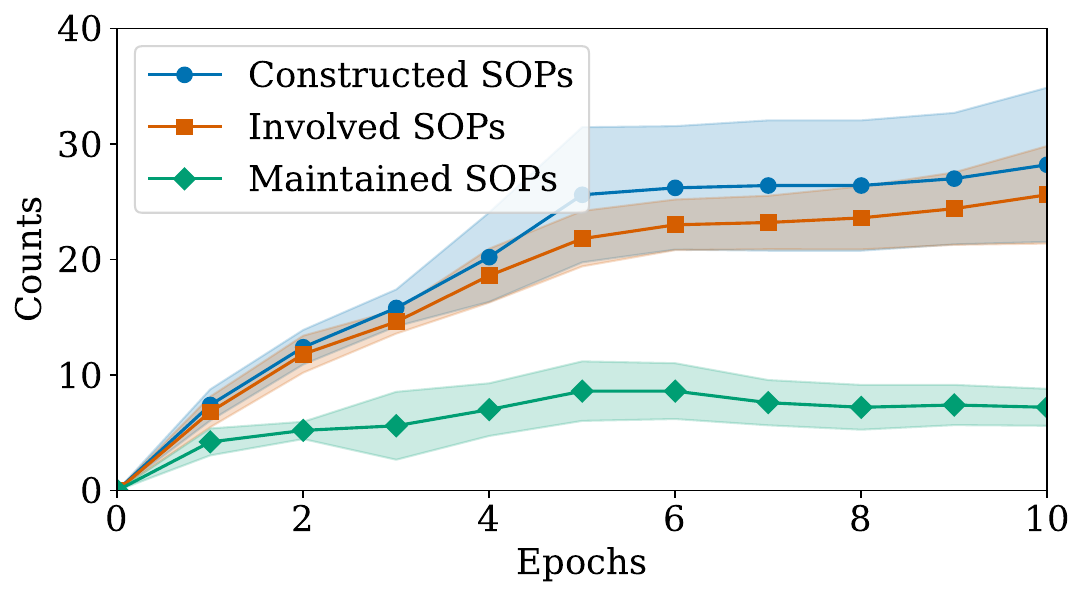}
  \vspace{-0.08in}
  \caption{Trends of constructed, involved, and maintained SOPs across training epochs in ACEBench.}
  \label{fig:toolsetcount}
\end{figure}

In this section, we supplement the description that EvoSOP maintains a condense but effective tool set at a low scale.
Figure~\ref{fig:toolsetcount} summarizes the macro evolution of the toolset's composition.
During the initial phase (Epochs 1 to 5), the toolset size undergoes rapid expansion as the \textsc{Constructor} explores various action patterns from the training trajectories. However, as the optimization loop proceeds, the growth curve flattens and eventually enters a dynamic equilibrium where new merged SOPs are only added if they offer significant utility gains over existing ones. 
Notably, \textsc{EvoSOP} maintains the toolset on a compact scale (typically fewer than 10 SOPs) while achieving a task success rate of approximately 80\%. Such stabilization mirrors the behavior of a decaying learning rate in traditional optimization, where the system converges to a minimal yet powerful set of SOPs. 

\section{Prompt Engineering in \textsc{EvoSOP}}
\label{app:prompts}

In this section, we summarize our prompt-engineering strategies used to implement the proposed LLM-based modules, namely \textsc{Constructor}, \textsc{Merger}, and \textsc{Reviewer}. In general, each component is treated as an independent agent. The prompt for each agent is carefully designed to provide a standardized description of its role, primary objective, operating paradigm, action boundaries, output requirements, and necessary few-shot examples.

We take the two \textsc{Constructor} functionalities, namely log analysis and SOP functionalization, as representative examples since they include structured content generation and coding. The other components and modules follow a similar prompt design logic. The following code block shows the prompt used for log analysis in \textsc{Constructor}.
\begin{minted}[fontsize=\footnotesize,breaklines,frame=single]{markdown}
## Identity
You are a specialized agent who is good at text extraction and analyzing. You are designed to summarize and extract some reusable and meaningful consecutive `tool_call`s from the provided action trajectories, forming a Standard Operation Process (SOP).

## Core Mission.
Your primary purpose is to summarize and extract the given workflows, induce reusable and meaningful consecutive `tool_call`s as an SOP.
There possibly contains more than one SOPs, or no SOP. Rewrite the SOPs which you think that satisfy the following rules, possibly multiple, one, or none.

### Operating Paradigm
You are given an action trajectory of a completed task containing `tool_call`s only, in the form of .json, the given arguments, and the name of that specific tool.
Some `tool_call`s have identifiable features which indicates that their functionality may have strong inter connections for combining them as an SOP.
Your paradigm of inducing SOPs are listed as follows:
1. **Consecutive `tool_call`**: This indicates that the `tool_call`s are executed consecutively.
2. **Separated `tool_call` Message Number**: This indicates that the `tool_call` are closely connected, but not executed consecutively. For example, `tool_call` A write down a python file, and encounters some error with consecutive `tool_call` B, but `tool_call` C is more general and successfully runs A without errors.
3. **Similar `tool_call` Results and Following Input Argument**: This indicates that two `tool_call`s are strongly connected, as the result of one `tool_call` is immediately used in another one's input.
4. **Meaningful Combination**: Some consecutive `tool_call`s combination are of great reality meanings, as `search` plusing `write_file` means search something and directly write this to some designated file.
5. **Frequently Usage**: Some consecutive `tool_call`s combination are of frequent occurrence, which implys that they can be modularized.
 
### Important Constraints and Hints
1. As is stated above, the **consecutive** `tool_call` means the they should be executed as close as possible. Do not induce `tool_call`s too far, making the span too big. 
2. The tools in induced SOP should have strong inter connection. You should carefully check the connection between `tool_call`s by examining the `tool_call` arguments and results. DO NOT INTEGRATE IRRELEVANT TOOL COMBINATIONS TO COMPLICATE THE RESULT.
3. The induced SOP should containing **at least** 2 `tool_call` actions to make the induced SOP be concise and not too simple.
4. The induced SOP should containing **no more than** 5 `tool_call` actions to make the induced SOP be general and scalable to be applied to other similar tasks.
5. Consecutive file operations have a higher probability of being combined together. You may need to consider deeper about the returned value of some `tool_call`, for the input parameters of the next `tool_call`.
6. Focus more on the workflow revealed by the logs. Do not care too much about the content in 'text' field if it is extremely long.
7. Except the required output in the following output guidance, DO NOT OUTPUT ANYTHING ELSE.

## Output Guidance
The output should just be a serialized python list object.
The list object may contain one valuable induced SOP, and the SOP should also be a list, containing the corresponding `message_number` in the `tool_call` logs.
If you think there is not valuable combination, output an empty list.

### Example

**Example Input:**

{example_trajectory}

**Example Output**

{example_tool_call_ids}

## Input Logs
The following contents are the input logs for you to process. Try your best to finish your task.
\end{minted}

The following code block shows the prompt of SOP functionalization in \textsc{Construtor}.
The ``function\_guidance'' should be a guidance and limitation of constructing SOPs, which could vary among different environments and datasets.

\begin{minted}[fontsize=\footnotesize,breaklines,frame=single]{markdown}

## Identity
You are a specialized agent who is good at software engineering. You are designed to rewrite the consecutive `tool_call`s to a standard operating procecure (SOP) from the provided action trajectories.

## Core Mission
Your primary purpose is to extract the current workflows, induce the `tool_call`s in the given logs to a well-bounded SOP, and rewrite the induced SOP in the form of a new tool function for future usage.

### Operation Paradigm
You are given an action trajectory of a completed task containing `tool_call`s only, in the form of .json, with the message number, the given arguments, and the name of that specific tool.
You are also given the docstring of all the involved `tool_call`s, which will help you conclude some implicit requirements or restraints about the tool function.

### Important Constraints and Hints
1. Consecutive file operations have a higher probability of being combined together. You may need to process the returned value of some `tool_call`, for the input parameters of the next `tool_call`. 
2. Focus more on the workflow revealed by the logs. Do not care too much about the content in 'text' field with repeated patterns.
3. Pay special attention to the docstrings of the input parameters of the `tool_call`s components, and reveal any special pattern if the input parameters are directly passed to these `tool_call`s. 
4. You should include all of the used `tool_call` names (in order if possible) in the docstring. 
5. Except the code, DO NOT OUTPUT ANY OTHER THINGS.

## Output Function Guidance

{function_guidance}

## Input Logs
The following contents are the input messages for you to process.

{trajectories}

The following contents are the docstrings of all the involved `tool_call`s.

{tool_docstrings}

\end{minted}

\section{An example of the synthesized SOP}
\label{code:sop_example}
In this section, we provide the source code of a representative SOP generated by \textsc{EvoSOP}. This SOP function contains four separate tool calls and a branching control mechanism that determines whether the message has been successfully sent directly, as well as the actions taken by the SOP in each case.

\begin{figure*}[h]
\begin{minted}[fontsize=\footnotesize, breaklines, frame=single, tabsize=2, linenos]{python}
import traceback
from typing import Any, Dict

def send_message_workflow(self, message: str, receiver_name: str, sender_name: str) -> Dict[str, Any]:
    """
    This function performs the following steps:
    1. Attempts to send a message from one user to another.
    2. If sending fails due to memory issues, it retrieves the latest message ID.
    3. Deletes the latest message based on its ID.
    4. Retries sending the message.
    
    Args:
        self: Instance of the current class.
        message (str): The message content to send.
        receiver_name (str): Name of the receiver.
        sender_name (str): Name of the sender.

    Returns:
        dict: A dictionary containing:
            - "status" (bool): True if the operation succeeds, False otherwise.
            - "content" (dict): Contains results of message sending, and deletion if applicable.
                - "send_status_initial": Result of the first send attempt.
                - "latest_message_id": ID of the latest message retrieved (optional).
                - "delete_status": Result of the delete operation (optional).
                - "send_status_final": Result of the retry send operation.
    """
    try:
        # Step 1: Attempt to send the message initially
        send_result_initial = self.send_message(message=message, receiver_name=receiver_name, sender_name=sender_name)

        # Check if sending failed due to memory issues
        if not send_result_initial['status']:
            # Step 2: Retrieve the latest message ID
            latest_message_result = self.get_latest_message_id()
            latest_message_id = latest_message_result.get('message_id')

            # Step 3: Delete the latest message based on its ID
            delete_result = self.delete_message(message_id=latest_message_id)

            # Step 4: Retry sending the message
            send_result_final = self.send_message(message=message, receiver_name=receiver_name, sender_name=sender_name)
            
            return {
                "status": send_result_final['status'],
                "content": {
                    "send_status_initial": send_result_initial,
                    "latest_message_id": latest_message_id,
                    "delete_status": delete_result,
                    "send_status_final": send_result_final
                }
            }
        
        # Message was sent successfully in the first attempt
        return {
            "status": send_result_initial['status'],
            "content": {
                "send_status_initial": send_result_initial
            }
        }
    
    except Exception as e:
        return {
            "status": False,
            "content": {
                "error": str(e),
                "traceback": traceback.format_exc()
            }
        }
\end{minted}
\end{figure*}

\end{document}